%% file: main.tex
\documentclass[twoside]{article}

\usepackage[accepted]{aistats2022}
\usepackage{url}
\usepackage{natbib}
\usepackage{graphicx}
\usepackage{booktabs}
\usepackage{enumitem}
\usepackage{subcaption}
\usepackage{lipsum}
\usepackage{cuted}
\usepackage{float}
\usepackage{dsfont}
\usepackage[colorlinks]{hyperref}
\hypersetup{
	citecolor = {blue}
}

\input{math_commands.tex}

\newcommand{\X}{\displaystyle \mX}

\newcommand{\zl}{\displaystyle \mZ_l}
\newcommand{\zt}{{\mathbf{z}}_t}
\newcommand{\zg}{{\mathbf{z}}_g}
\usepackage[bottom]{footmisc}

\begin{document}


\runningauthor{Tonekaboni, Li, Ar{\i}k ,Goldenberg, Pfister}

\twocolumn[

\aistatstitle{Decoupling Local and Global Representations of Time Series }

\aistatsauthor{Sana Tonekaboni \footnotemark[1] \And Chun-Liang Li \And  Sercan \"{O}. Ar{\i}k}% \And Anna Goldenberg \And Tomas Pfister }
\aistatsaddress{ University of Toronto \And  Google \And Google}
\aistatsauthor{Anna Goldenberg \And Tomas Pfister }
\aistatsaddress{ University of Toronto \And  Google} ]

% \And University of Toronto \And  Google} ]

\footnotetext{\footnotemark[1]Work done while at Google}

\renewcommand{\thefootnote}{\arabic{footnote}}

\begin{abstract}
Real-world time series data are often generated from several sources of variation. 
Learning representations that capture the factors contributing to this variability enables a better understanding of the data via its underlying generative process and  improves performance on downstream machine learning tasks. 
This paper proposes a novel generative approach for learning representations for the global and local factors of variation in time series. 
The local representation of each sample models non-stationarity over time with a stochastic process prior, and the global representation of the sample encodes the time-independent characteristics. 
To encourage decoupling between the representations, we introduce counterfactual regularization that minimizes the mutual information between the two variables. 
In experiments, we demonstrate successful recovery of the true local and global variability factors on simulated data, and show that representations learned using our method yield superior performance on downstream tasks on real-world datasets. 
We believe that the proposed way of defining representations is beneficial for data modelling and yields better insights into the complexity of real-world data.\footnote{The implementation of our framework and the experiments can be found at: \url{https://github.com/googleinterns/local_global_ts_representation}}
\end{abstract}

\input{introduction}

\input{related_work}

\input{method}

\input{experiments}

\input{conclusion}

% \clearpage
% \bibliographystyle{natbib}
% {apalike}
\bibliography{ref}

\clearpage
\input{appendix}

\thispagestyle{empty}

% For one-column format, uncomment the following:
% \onecolumn \makesupplementtitle
% For two-column format, uncomment the following:
% \twocolumn[ \makesupplementtitle ]

\end{document}



% Supplementary material: To improve readability, you must use a single-column format for the supplementary material.
\onecolumn
\aistatstitle{SUPPLEMENTARY MATERIAL}

% \section{FORMATTING INSTRUCTIONS}

% To prepare a supplementary pdf file, we ask the authors to use \texttt{aistats2022.sty} as a style file and to follow the same formatting instructions as in the main paper.
% The only difference is that the supplementary material must be in a \emph{single-column} format.
% You can use \texttt{supplement.tex} in our starter pack as a starting point, or append the supplementary content to the main paper and split the final PDF into two separate files.

% Note that reviewers are under no obligation to examine your supplementary material.

\subsection{Datasets} \label{app:dataset}

\subsection{Experiment Setup} \label{app:exp}

Talk about all the hyper parameters that were chosen in each experiment ($\lambda$, $\beta$, model architectures, ...)

\subsection{Simulated Dataset} \label{app:sim_data}

\subsection{Supplementary Plots} \label{app:extra_plot}

\begin{figure}
    \centering
    \includegraphics[scale=0.35]{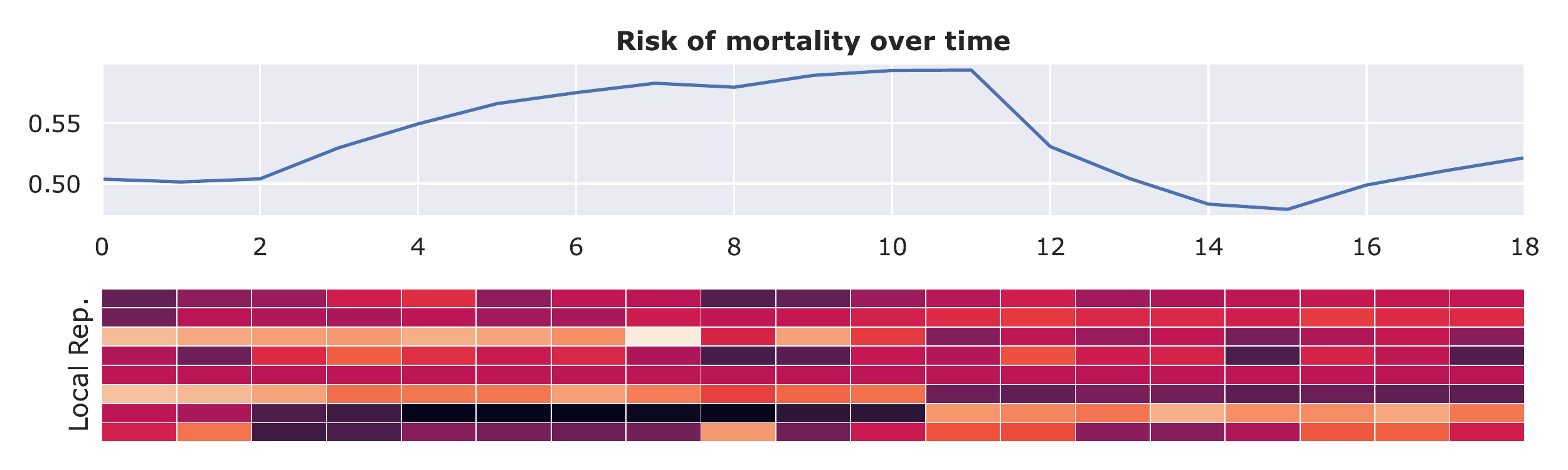}
    \caption{Risk of mortality estimation over time, based on the local and global representations. }
    \label{fig:mortality_risk}
\end{figure}

%% file: math_commands.tex
\usepackage{amsmath,amsfonts,bm}

\def\eqref#1{equation~\ref{#1}}
\def\1{\bm{1}}

\def\rvz{{\mathbf{z}}}

\def\mX{{\bm{X}}}

\def\mZ{{\bm{Z}}}

\DeclareMathAlphabet{\mathsfit}{\encodingdefault}{\sfdefault}{m}{sl}
\SetMathAlphabet{\mathsfit}{bold}{\encodingdefault}{\sfdefault}{bx}{n}

\newcommand{\E}{\mathbb{E}}

\newcommand{\KL}{D_{\mathrm{KL}}}

\newcommand{\Cov}{\mathrm{Cov}}

%% file: introduction.tex
\section{Introduction}

\begin{figure*}[h!]
     \centering
    \includegraphics[scale=0.165]{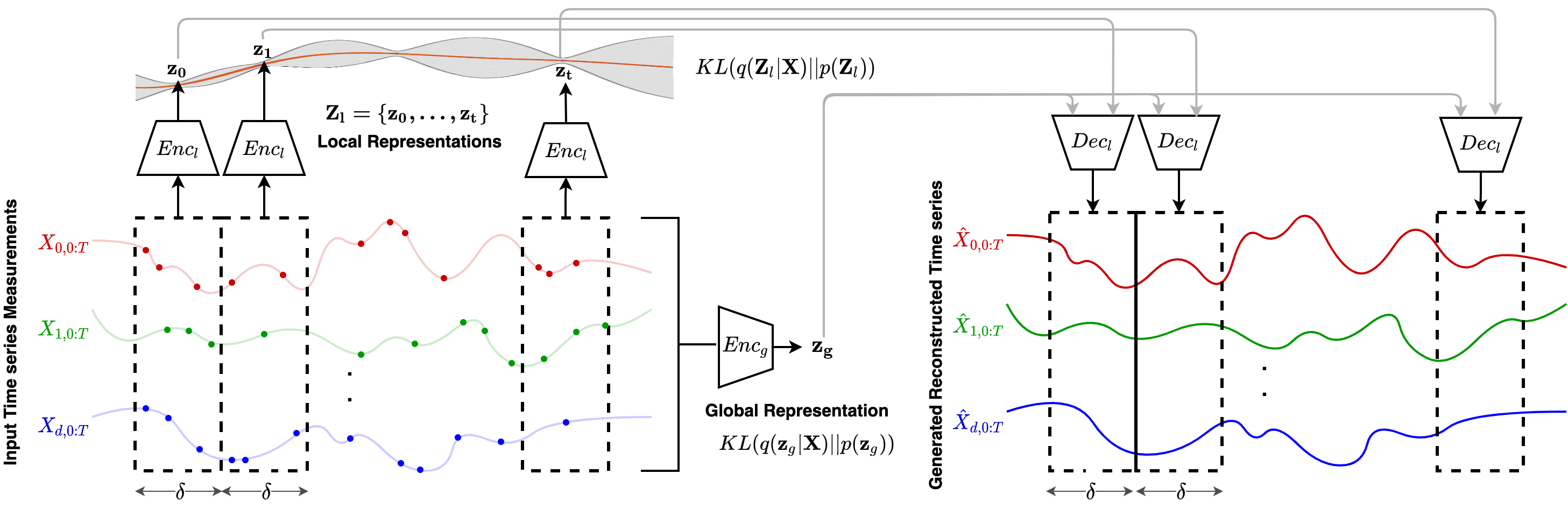}
    \caption{Overview of our method for learning global and local representations. 
    The proposed method learns the distribution of the local representations of windows of samples over time using the local encoder $Enc_l$. 
    The global encoder $Enc_g$ models the posterior of the global representation, and the decoder $Dec$ generates the time series using samples from the posterior of the local and global representations. }
    \label{fig:overview}
\end{figure*}

Learning high-quality representations for complex data such as time series helps better understanding the underlying generative process of the data \citep{ghahramani2015probabilistic} and  has a substantial impact on the performance of downstream machine learning (ML) tasks \citep{bengio2013representation}. 
Condensing information into expressive lower-dimensional representations can improve interpretability \citep{bai2018interpretable} and lead to better generalization and domain adaptation \citep{chen2012marginalized}. 
One challenge for downstream ML tasks for time-series data is that collecting labels can be expensive, as most real-world tasks may involve human expert domain knowledge. Also, the underlying state of the data may evolve, rendering human understanding of the data more difficult. 
Unsupervised representation learning would be of extortionate value in such cases, providing a powerful tool to uncover the underlying state and summarize the complex data into informative general-purpose representations.

Observational time series are often generated from different underlying factors of variation that can be identified through the representations. 
Some of these factors present the global attributes of a sample and are unique to each individual, while others are descriptive of the variations of the underlying state over time. 
For instance, consider a medical diagnosis application: the observed patient physiological signals are a product of the individual’s attributes such as gender, age and pre-existing medical conditions, and the medical treatment they received over time. 
These hidden factors could influence the time series in different ways. 
Uncovering and decoupling these factors is a fundamental challenge in representation learning that will lead to substantial improvement in data understanding as well as the performance of ML models that utilize the learned representations. 

In this paper, we introduce an unsupervised representation learning method for time series that decouples the global and local properties of the signal into separate representations.
Our method is based on generative modelling and assumes that each time series sample is generated from two underlying factors: a global and local representation. 
The global representations are unique to each sample and represent static time-independent characteristics. 
The local representations encode the dynamic underlying state of windows of the time series sample as they evolve over time, taking into account the non-stationarity of the samples. 
We use variational approximation to model the posterior of the two sets of representations and model the temporal dynamics using a prior sampled from a Gaussian Process (GP).
To ensure that the two sets of representations are decoupled and they model distinct data characteristics, we introduce a counterfactual regularization with the goal of mutual information minimization to disentangle the representations. 
Fig. \ref{fig:overview} overviews our framework.
Decoupling local and global representations using our proposed method provides the following important benefits:
\begin{enumerate}
    \item Exploiting global patterns in the data and coupling them with local variations serves as an inductive bias and improves the downstream tasks from time series {\bf{forecasting}} performance to {\bf classification} accuracy. 
    \item By disentangling the factors of variation, we can represent the underlying information in a more {\bf{efficient}} way by encoding the necessary information for target tasks with more compact representations.  
    \item Knowing the various factors of variation in generation of a time series helps identify and disentangle the underlying explanatory factors and results in more {\bf{interpretable}} representations. 
    \item Having this information also allows one or the other representations to be used flexibly for downstream tasks depending on which property is appropriate for a specific use case. 
    For instance, global representations can help us {\bf{identify subgroups}} with similar properties.
    \item Using our approach, we can {\bf{generate counterfactual instances}} of data with controlled factors of variation, utilizing the global and local representations. 
    This enables generating different manifestations of events like disease progression in individuals with different characteristics. 
\end{enumerate}

% Our results show that our method generates informative and low-dimensional representations that can identify the global and local patterns of the data (demonstrated through simulations) and can be used to improve downstream tasks. Learning the global and local representations independently improves modeling of the signal and also helps with better understanding of the data's underlying explanatory factors. 
%Impact statement?

%% file: related_work.tex
\section{Related Work}
\label{sec:related_work}

The performance of many ML models relies heavily on the quality of the data representations \citep{Bengio+chapter2007}. 
This applies to all data types, but it is vital for complex ones such as time series that can be high-dimensional, high-frequency and non-stationary \citep{yang200610, langkvist2014review}. 
Due to the difficulties in labelling time-series data, unsupervised approaches are often preferred in such settings. They include different categories of methods that are based on reconstruction \citep{yuan2019wave2vec, fortuin2018som, fortuin2020gp, chorowski2019unsupervised}, clustering \citep{ma2019learning, lei2019similarity}, contrastive objectives \citep{oord2018representation, franceschi2019unsupervised, hyvarinen2019nonlinear, tonekaboni2020unsupervised, hyvarinen2016unsupervised}, and others.

While all the methods above have been shown to successfully encode the informative parts of the signal in a low dimensional representation, improving the interpretability of the representations is still an active area of research. 
An effort in this direction is disentangling the dimensions of the encoding, which enables representing different factors of variation in independent dimensions. 
Earlier approaches learn disentangled representations using supervised data \citep{yang2015weakly, tenenbaum2000separating}, while more recent methods provide unsupervised solutions to tackle this problem using adversarial training \citep{chen2016infogan, kim2018disentangling, kumar2018variational} or regularization of the data distribution \citep{dezfouli2019disentangled}. 
However, associating these factors with interpretable notions in the data domain remains challenging. 
A different line of work focuses on decoupling global and local representations of samples into separate representations (or dimensions of representation). 
This idea has been explored for visual data to separate the factors of variation associated with the labels from other sources of variability. 
\citet{mathieu2016disentangling} use a conditional generative model with adversarial regularization, while \citet{ma2020decoupling} learn decoupled representations of global and local information of images relying on empirical characteristics of VAE and flow models.

For time series, VAE-based methods have been used to disentangle dynamic and static factors of representations for video and speech data. FHVAE \citep{hsu2017unsupervised} uses a hierarchical VAE model to learn \emph{sequence-dependent variables} (corresponding to the speaker factor) and \emph{sequence-independent variables} (corresponding to the linguistic factors) in modeling speech, but it does not explicitly encourage disentanglement. DSVAE \citep{yingzhen2018disentangled} formalizes the idea of disentangled representation by explicitly factorizing the posterior to model the static and dynamic factors. Other methods have been proposed to augment DSVAE by explicitly enforcing disentanglement in their objective function. S3VAE \citep{zhu2020s3vae} introduces additional loss terms that encourage the invariance of the learnt representations by permuting the frame and leveraging external supervision for the motion labels. An additional regularization is imposed by minimizing the mutual information between the two sets of representations. Similar regularizations are used in C-DSVAE \citep{bai2021contrastively} using contrastive estimation of mutual information. Note that for all the above methods, the focus is on the generated samples and not the quality nor interpretability of the representations. Therefore, the dynamic local representations are not designed to summarize the information of the time series samples over time.
Other efforts for disentangling global and local representations are designed to improve specific downstream modeling tasks. 
For instance, \citet{sen2019think} leverage both local and global patterns to improve forecasting based on matrix factorization techniques. Similar ideas have also been introduced to improve forecasting performance \citep{wang2019deep, nguyen2021temporal}.
\citet{schulam2015framework} learn population and sub-population parameters to model personalized disease trajectories. 
% In this work, our goal is to learn decoupled global and local representations that best explain the underlying generative processes.

%% file: method.tex
\section{Method}

%We model the given time series data with the assumption that they are generated from two independent sources of variability. 
In this section, we present the notations used throughout this paper, followed by the problem definition and description of the method. 
% The first represents the global properties, which are the unique characteristics of a sample that remain the same over time. Second, is the local representation that determines the underlying state of a sample over time and can capture the potential non-stationarity. 
% For instance in a clinical signal example, the global representation can encode the pre-existing condition of a patient and the local representations will show the evolution of the individual's health state or disease progression.

%Similar assumptions have been made for other data types, such as images where the global representations are often referred to classes of images, and the local representations model other variations within each class such as the object orientation etc. \citep{mathieu2016disentangling, ma2020decoupling}.

%%%%%%%%%%%%%%%%%%%%%%%%%%%%%%%%%%%%%%%%%%%%%%
%%%%%%%%%%%%%%%% Notation %%%%%%%%%%%%%%%%%%%%
%%%%%%%%%%%%%%%%%%%%%%%%%%%%%%%%%%%%%%%%%%%%%%

\subsection{Notation}
Let $\X^{(i)} \in \mathbf{R}^{d\times T}$ be a multivariate time series sample ($i\in N$), with $d$ input features and $T$ measurements over time. 
Each time series sample is generated from two latent variables $\zg$ and $\zl$. 
The global representation $\zg^{(i)}  \in \mathbf{R}^{d_g}$ is a vector of size $d_g$ that represents the global properties of sample $i$. 
The local representation of sample $i$,
$\zl^{(i)}=[\rvz_0^{(i)}; \rvz_1^{(i)}; \dots; \rvz_t^{(i)}]$, is composed of a set of vector representations of non-overlapping windows $[\X^{(i)}_{0}; \X^{(i)}_{1}; \dots; \X^{(i)}_{t}]$ of time series $\X^{(i)}$. 
Each $\zt^{(i)} \in \mathbf{R}^{d_l}$ is the representation for a window of time series $\X^{(i)}_t\in \mathbf{R}^{d\times \delta}$ of length $\delta$ that encodes information for all features within the window.
The windows split the sample into $\lceil \frac{T}{\delta} \rceil$ consecutive parts as shown in Fig. \ref{fig:overview}. 
The size of the global representations and the local representation is determined by $d_g$ and $d_l$, respectively. 
In order to handle missing measurements, each sample has a mask channel $M^{(i)} \in \mathbf{R}^{d\times T}$ to indicate which data points are measured and which ones are missing. 
Irregularly-sampled time series can also be converted into regularly-sampled signals with the additional mask channel to indicate the measurements.
For simplicity, we drop the sample index $i$ in the rest of the paper.

\begin{figure}[!htb]
    % \centering
     \includegraphics[scale=0.19]{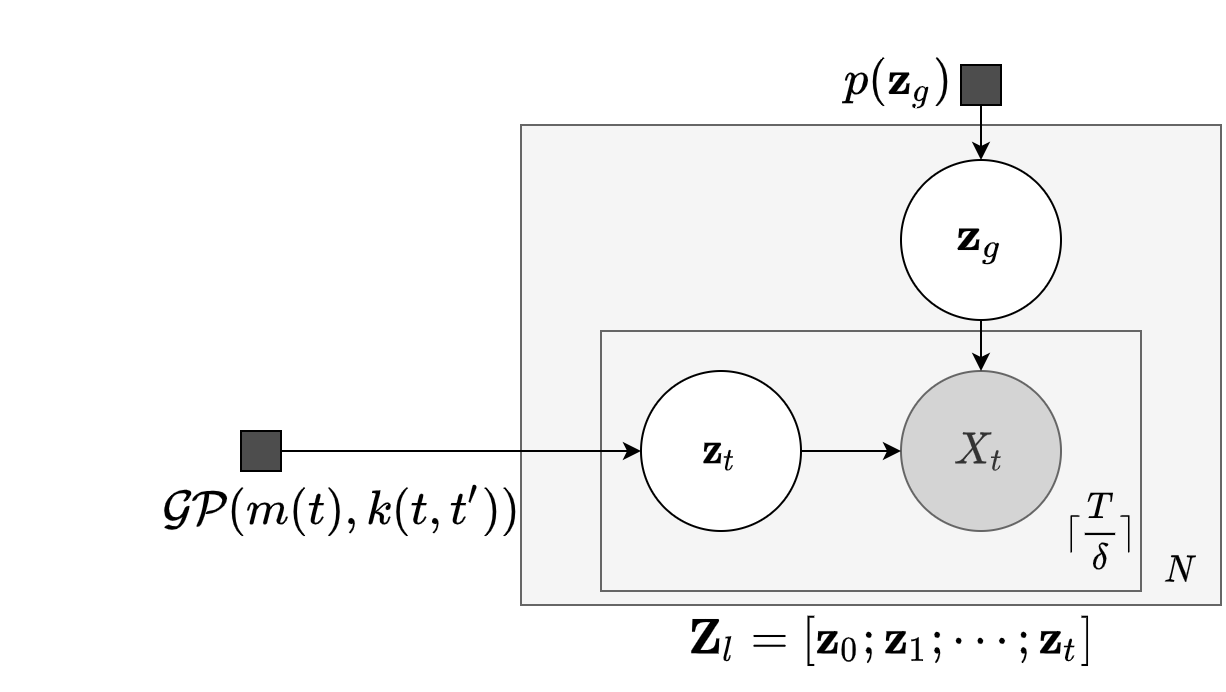}
    \caption{Graphical model of the generative process. 
    Each window of time series $\X_t$ is generated from the global representation of the samples $\zg$ and the local representation of the window $\zt$. 
    The sample $\X$ is composed of the series of consecutive windows $\X_t$, and the pset of all local representations $\zl =[\rvz_0; \rvz_1; \dots; \rvz_t]$ are sampled from a Gaussian Process prior.}
    \label{fig:graphical_model}
\end{figure}

Our probabilistic data generation mechanism assumption is shown in Fig. \ref{fig:graphical_model}. 
We model the conditional likelihood distribution of the data as follows: $\X \sim p( \X |\zl, \zg)$. $\X_t$ in Fig. \ref{fig:graphical_model} represents a window of time series, and $\rvz_t$ is the local representation for that window. 
The local representation of the windows can change over time as the underlying state of the time series changes. 
The dependencies of these local representations are modeled using a prior sampled from $\mathcal{GP}(m(t),k(t,t^\prime))$. 
The global representation $\zg$ is the same for all windows within a sample, and its prior is modelled as a Normal Gaussian distribution.

%%%%%%%%%%%%%%%%%%%%%%%%%%%%%%%%%%%%%%%%%%%%%%
%%%%%%%%%%%%% Inference Model %%%%%%%%%%%%%%%%
%%%%%%%%%%%%%%%%%%%%%%%%%%%%%%%%%%%%%%%%%%%%%%

\subsection{Modeling Distributions}

As parts of the learning algorithm, we use variational approximations to model the following three distributions: 
\begin{enumerate}
    \item The conditional likelihood distribution of the time series sample $p( \X |\zl, \zg)$ conditioned on the local and global representations. We approximate this using a {\bf{Decoder}} model ($Dec(\zl, \zg)$).
    \item The posterior over the local representations $q(\zl|\X)$. We approximate this using the {\bf{Local Encoder}} ($Enc_l(\X)$). This encoder slices the time series into consecutive windows and approximates the joint distribution of all the local representations over time.
    \item The posterior distribution of the global representations $q(\zg|\X)$. The {\bf{Global Encoder}} ($Enc_g(\X)$) approximates the parameters of this conditional distribution. The encoder can input the entire time series, or any part of it, for estimating the global representations. This is particularly useful in the presence of missing data and allows using the part of the signal with fewer missing observations. To ensure robustness and to encourage the global representation to be constant throughout a sample, the encoder is trained on random sub-windows of the sample. 
\end{enumerate}

The local representation should model the temporal behaviour and the underlying time-varying states. The local encoder learns the representation of consecutive windows independently; however, we cannot assume that these representations are $i.i.d$. \citet{casale2018gaussian} show that for video data, by accounting for temporal covariance between the representations, we can learn more general and informative encodings. Similar to ideas presented in \citet{casale2018gaussian} and \citet{fortuin2020gp}, we impose the temporal dependencies between the local representations using a prior sampled from a Gaussian Process (GP). GP is a non-parametric Bayesian method well-suited for modelling temporal data that helps with robust modelling even in the presence of uncertainty \citep{roberts2013gaussian}. We model each dimension of the local representations (indexed by $j \in d_l$) independently over time, using unique GPs with different kernel functions. The intuition behind this choice is to decompose the latent representations into dimensions with unique temporal behaviours that are characterized by the covariance structure. E.g., a dimension with a periodic kernel can model the seasonality of the underlying state of the signal. This is an important property since the local representations should model the non-stationarity in the samples. The local encoder $Enc_l(\X)$ approximates the posterior of each dimension using a multivariate Gaussian distribution as shown in Eq. \ref{eq:variational_dist}.
\begin{equation}
    q(Z_{0:T, j}|\X) = \mathcal{N}(Z_{0:T,j};\mu_{j}, \Sigma_{j}).
    \label{eq:variational_dist}
\end{equation}
Following \citet{fortuin2020gp}, the precision matrix of the covariance $\Sigma_j^{-1}$, is parameterized as a product of bi-diagonal matrices (Eq. \ref{eq:covariance}), where $B_j$ is an upper triangular band matrix that guarantees positive definitiveness and symmetry in $\Sigma_j^{-1}$. With this sparse estimation of the precision matrix, sampling from the posterior becomes more efficient \citep{mallik2001inverse} and the estimated covariance can still be dense and model long-range dependencies in time.

\begin{equation}
    \Sigma_j^{-1} = B_j^TB_j,  
    \{B_j\}_{tt^\prime} =
    \begin{cases}
      b^j_{tt^\prime} & \text{if } {t}^\prime \in \{t, t+1\},\\
      0, & \text{otherwise.}
     \end{cases}
    \label{eq:covariance}
\end{equation}

With the parametric approximations for the distributions, we can use the $ELBO$ objective in Eq. \ref{eq: nll_loss} to train the models.

\begin{equation}
\begin{split}
    \mathcal{L} & = -\displaystyle  \E_{\zl,\zg} [\log(p(\X|\zl, \zg))]  \\     & + \beta [\displaystyle \KL (q(\zl|\X) \Vert p(\zl)) \\  &+ \displaystyle \KL (q(\zg|\X) \Vert p(\zg))]
\end{split}
\label{eq: nll_loss}
\end{equation}

The log-likelihood term ensures the generated signals are realistic, and the $\displaystyle \KL$ divergence terms minimize the distance between the estimated distributions and their priors. As mentioned, the prior over the global representations $p(\zg)$ is assumed to be a standard Gaussian ($\mathcal{N}(0,1)$) and the prior over the local representations $p(\zl)$ is a zero-mean GP prior defined over time
% Essentially, for each dimension of the local representations $\kappa \in d_l$, the prior is defined by a zero-mean GP and a kernel function for the covariance 
($\mathcal{GP}(0,k_j(t,t^\prime))$). 
We assume different kernel functions and parameters (i.e. the length scale) for different dimensions of the latent representations to model various dynamics at multiple time scales. Our framework is compatible with many kernel structures, including but not limited to RBF, Cauchy, and periodic. A list of kernel functions used in our experiments is presented in the Appendix \ref{app:exp}.
Note, the negative log-likelihood is only estimated for the observed measurements to account for the missing values.

Eq. \ref{eq: nll_loss}, however, does not guarantee all the properties that we expect from the representations. Nothing would prevent all information to flow through the local representation $\zl$, which has a higher encoding capacity. This means that the model can easily converge to a solution where all information about the global behaviour of the signal is encoded in the local representations. As a result, $\zg$ would become random noise ignored by the decoder and the local representation would no longer represent underlying states, independent of sample variabilities. To address this, we introduce the counterfactual regularization term of the loss, described in the next section.

%%%%%%%%%%%%%%%%%%%%%%%%%%%%%%%%%%%%%%%%%%%%%%
%%%%%%%%%%% Counterfactual Reg. %%%%%%%%%%%%%%
%%%%%%%%%%%%%%%%%%%%%%%%%%%%%%%%%%%%%%%%%%%%%%

\subsection{Counterfactual Regularization}

\begin{figure}[h!]
    \centering
    \includegraphics[scale=0.2]{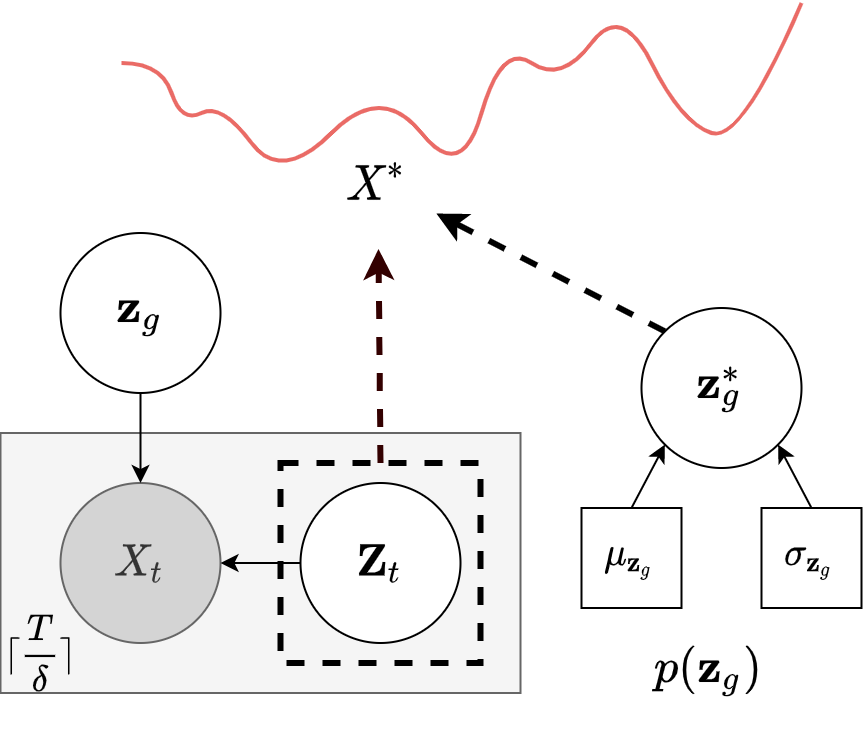}
    \caption{The counterfactual sample $\X^*$ generation process for the regularization term. }
    \label{fig:counter_factual}
\end{figure}

The issue of information flowing through one set of representations can be prevented if we have labels available for the global sources of variations \citep{reed2015deep, klys2018learning}. However, in practice, they are rarely available, and the underlying factors are often unknown. We propose a counterfactual regularization,  encouraging $\zg$ to be informative and the global behaviours to be only encoded in the global representation. This means that variation in the local representation should not change the global identity of the time series and vice versa. For each sample $\X^{(i)}$ during training, we generate a counterfactual sample $\X^*$ with the local representation of $\X^{(i)}$ ($\zl^*=\zl^{(i)}$), and a random global representations sampled from the prior ($\zg^* \sim p(\zg)$), as explained in Fig. \ref{fig:counter_factual}. 
% For each sample $\X^{(i)}$ during training, we generate a counterfactual sample that has the local properties of $\X^{(i)}$, and the global properties of another sample $\X^{(j)}$ (as explained with the example in Figure \ref{fig:counter_factual}). 

Ideally, this generated counterfactual sample $\X^*$ would have no signs of the global properties of $\X^{(i)}$. If the two representations are independent, $\zl^{(i)}$ should not contain any information about $\zg^{(i)}$; therefore, $\zg^{(i)}$ should have low likelihood under the estimated posterior distribution of the global representations, conditioned on the counterfactual sample $q(\zg|\X^*)$. Using the global encoder, we estimate this posterior distribution ($Enc_g(X^*)$) and encourage the likelihood ratio for $\zg$ to $\zg^*$ to be low. Our proposed regularization term is as follows:
\begin{equation}
    % \mathcal{L}_{reg} = -\log(\frac{-(d(\zg^{(i)}, Enc_g(\X^{*(i)}))-1)}{2})
    \mathcal{L}_{reg} = \E_{\zg, \zl} \frac{q(\zg|\X^*)}{q(\zg^*|\X^*)},
    \label{eq:cf_req}
\end{equation}
As a result, the final objective  becomes:
% We enforce this property by maximizing the distance between $Enc_g(\X^{*(i)})$ and $\zg^{(i)}$, as formulated in Equation \ref{eq: cf_req}. $d(\cdot)$ is a similarity metric to measure the distance between the estimated global representation vectors, and the $\log$ is to break the cost linearity. We have chosen the cosine similarity function in our applications. 
\begin{equation}
\begin{split}
\mathcal{L} = & -\displaystyle  \E_{\zl,\zg}%\sim q(\zl|\X)} 
[\log(p(\X|\zl, \zg))] \\ & + \beta [\displaystyle \KL ( q(\zl|\X) \Vert p(\zl) )\\
& + \displaystyle \KL ( q(\zg|\X) \Vert \mathcal{N}(0,1)) )] + \lambda \mathcal{L}_{reg}.
\end{split}
\label{eq:objective}
\end{equation}

% \begin{equation}
%     \mathcal{L} = -\displaystyle  \E_{\zl\sim q(\zl|\X)} [log(p(\X|\zl, \zg))] +  \beta \displaystyle \KL ( q(\zl|\X) \Vert p(\zl) ) + \lambda \mathcal{L}_{reg}
%     \label{eq:objective}
% \end{equation}

\paragraph{Counterfactual regularization and disentanglement.} An essential role of counterfactual regularization is encouraging independence implicitly. 
One way to achieve independence between global and local variables is through minimizing the mutual information $I(\zg, \zl)$ between the two variables. 
Following~\cite{moyer2018invariant}, the mutual information between the two sets of representations can be decomposed as:

\begin{equation} \label{eq:MI}
\begin{split}
I(\zg;\zl) & = I(\zg; \X) - I(\zg; \X | \zl) \\
 & = I(\zg; \X) {+} H(\zg|\X) {-} H(\zg|\zl) \\
 & = I(\zg; \X) {+} H(\zg|\X) {+} \E_{\zg, \zl}\log p(\zg|\zl). \\
% & \approx I(\zg, \X) + c + \E_{\zg, \zl}\log q(\zg|\zl)
\end{split}
\end{equation}
The first two terms measure the information captured by the global representation $\zg$, which is also considered in the variational autoencoder objective (Eq. \ref{eq: nll_loss}). Minimizing $I(\zg;\zl)$ can therefore be done by minimizing $\E_{\zg, \zl}\log p(\zg|\zl)$. As we do not have access to the distribution $p(\zg|\zl)$, existing works~\citep{cheng2020club} use an additional network to construct the variational approximation.
Instead of introducing an additional network to approximate $p(\zg|\zl)$, which increases training complexity and computation time, we reuse the global encoder for counterfactual regularization for this approximation as follows. Since $\zg^*$ is sampled from the prior distribution independent of $\zl$, then $q(\zg|\zl)=q(\zg|\zl,\zg^*)\simeq q(\zg|\X^*)$ if the decoder $\zl,\zg^*\rightarrow \X^*$ preserves the information. This implies that minimizing~(\ref{eq:cf_req}) implicitly minimizes $I(\zg;\zl)$ and encourages decoupling between the representations.

%We use the counterfactual sample as a way to find a variational approximation of the conditional distribution $q(\zg|\zl)$. The global encoder predicts the posterior over the global representation, given the counterfactual sample $\X^*$, which is a function of $\zl$ and a random global variable $\zg^*$.

% \begin{equation} \label{eq:MI}
% \begin{split}
% H(\zg|\zl) & = \E_{\zg, \zl}[-\log p(\zg|\zl)] \\
%  & = \inf_{r(\zg|\zl)} \E_{\zg, \zl}[-\log r(\zg|\zl)]
% \end{split}
% \end{equation}

% This implies that another way of minimizing $I(\zg, \zl)$ is to optimize the conditional distribution $p(\zg|\zl)$ such that $r(\zg|\zl)$, the lowest entropy predictor of $\zg$ given $\zl$, has the highest entropy. This intuitively means that, a predictor should be inaccurate in predicting the global properties, just knowing the local representations, and this is exactly what the regularization is encouraging. $\X^{(i)}$ is a sample generated from $\zl^{(i)}$, but with a random global behaviour, and the predictor of $p(\zg^{(i)}|\zl^{(i)})$ is encouraged to perform badly.

%% file: experiments.tex
\section{Experiments}

\begin{table*}[!h]
\begin{center}
\begin{tabular}{llccccc}
 & & \multicolumn{2}{c}{ICU Mortality Prediction} & \multicolumn{1}{c}{Average Daily Rain Estimation} \\
\toprule
{\bf Model}  & {\bf Dimensions} &{\bf AUPRC} & {\bf AUROC}  & {\bf Mean Absolute Error} \\ 
\midrule
Our method & 8 $\times$ steps + 8 & {\bf0.365} $\pm$ 0.092  & {\bf0.752} $\pm$ 0.011 & 1.824 $\pm$ 0.001\\%{\bf 1.824} $\pm$ 0.002\\
Our method - no reg. & 8 $\times$ steps + 8 & 0.238 $\pm$ 0.026     & 0.672 $\pm$ 0.010 & 1.825 $\pm$ 0.001 \\
\midrule
GP-VAE & 8 $\times$ steps  &  0.266 $\pm$ 0.034  & 0.662 $\pm$ 0.036 & 1.824 $\pm$ 0.001\\
GP-VAE & 16 $\times$ steps  &  0.282 $\pm$ 0.086 & 0.699 $\pm$ 0.018 & 1.826 $\pm$ 0.001\\%1.834 $\pm$ 0.014 \\
VAE & 8 $\times$ steps &  0.157 $\pm$ 0.053 & 0.564 $\pm$ 0.044& 1.831 $\pm$ 0.005\\%1.842 $\pm$ 0.013 \\
VAE & 16 $\times$ steps &  0.118 $\pm$ 0.001 & 0.491 $\pm$ 0.037 & 1.840 $\pm$ 0.012\\%1.833 $\pm$ 0.011 \\
C-DSVAE & 8 $\times$ steps + 8 & 0.158 $\pm$ 0.005 	& 0.565 $\pm$ 0.007 & {\bf1.806} $\pm$ 0.012\\
\midrule
Supervised & NA     &  {\bf 0.446} $\pm$ 0.036 & {\bf 0.802} $\pm$ 0.043 & {\bf 0.079} $\pm$ 0.001\\
\bottomrule
\end{tabular}
\caption{Performance of different representation learning methods on downstream predictive tasks.}
\label{tb:downstream}
\end{center}
\end{table*}

\begin{table*}[!ht]
\begin{center}
\begin{tabular}{lccc}
& & \multicolumn{1}{c}{Air Quality} & \multicolumn{1}{c}{Physionet}\\
\toprule
{\bf Model}  & {\bf Dimensions}  & {\bf Accuracy} & {\bf Accuracy}
\\ \midrule
%Lambda=0.5 for air_quality and lambda=1.0 for physionet
Our method & 8  & {\bf 57.93} $\pm$ 3.53 &  {\bf 46.98} $\pm$ 3.04 \\
Our method - no reg. & 8   & 38.35 $\pm$ 2.67 & 32.54 $\pm$ 0.00\\
\midrule
GP-VAE & 8   & 36.73 $\pm$ 1.40 & 42.47 $\pm$ 2.02
\\%34.39 $\pm$ 0.45   & 1.32 $\pm$ 0.00
GP-VAE & 16      & 33.57 $\pm$ 1.50 & 44.67 $\pm$ 0.50 \\
VAE & 8  & 27.17 $\pm$ 0.03 & 34.71 $\pm$ 0.23\\
VAE & 16    & 31.20 $\pm$ 0.33 & 35.92 $\pm$ 0.38\\ 
C-DSVAE & 8  & 47.07 $\pm$ 1.20 & 32.54 $\pm$ 0.00\\
% \midrule
% TNC & 8  & 57.73 $\pm$ 3.80 & 44.84 $\pm$ 0.96\\
% CPC & 8  & 37.20 $\pm$ 0.27 & 43.36 $\pm$ 0.23\\
\midrule
Supervised &    NA  & {\bf 62.43} $\pm$ 0.54 & {\bf 62.00} $\pm$ 2.10\\
\bottomrule
\end{tabular}
\caption{Performance of different representation learning methods on identifying similar subgroups of data.}
\label{tab:global_classification}
\end{center}
\end{table*}

Evaluating unsupervised representation learning is challenging due to the lack of well-defined labels for the underlying representations and factors of variation. However, the representations' generalizability and informativeness can be assessed on different downstream tasks. We present a number of experiments to evaluate the performance of our method in comparison to the following benchmarks: (i) Variational Auto Encoder (VAE) as a standard unsupervised representation learning framework, (ii) GP-VAE \citep{fortuin2020gp}, (iii) C-DSVAE \citep{bai2021contrastively}, which separates dynamic and static representations \footnote{The implementation of this method for time series data doesn't have the augmentations proposed for video setting because such transformations (cropping, color distortion, etc.) are not defined for time series.}, and (iv) a model trained supervised for the task. 

All baselines are trained to learn representations for consecutive windows in the time series sample. For consistency and to allow comparison of performance results, the encoder and decoder architectures of different baselines are kept the same. 
One of the benefits of decoupling local and global representations is that we can condense the representation into fewer dimensions, especially since the global representation is unique throughout the sample. We have chosen different encoding sizes for our baselines to reflect this advantage better. The performance of all models is compared across multiple evaluation tests using two different time series datatset:

\begin{enumerate}[leftmargin=*]
    \item Physionet ICU Dataset \citep{goldberger2000physiobank}: A medical time-series dataset of records from 12,000 adult ICU stays in the Intensive Care Unit. The temporal measurements consist of physiological signals and different lab measurements. For each recording, there are general descriptors of the patient (age, type of ICU admission, etc. ) as well as labels indicating in-hospital mortality. We use such descriptors as approximates for the global properties of the signals.
    \item UCI Beijing Multi-site Air Quality Dataset \citep{zhang2017cautionary}: This dataset includes hourly measurements of multiple air pollutants from 12 nationally-controlled monitoring sites, collected over four years. The measurements are also matched with meteorological data from the nearest weather station. We partition the data such that each time series sample is the pollutant reading for parts of a particular month of the year.
\end{enumerate}

To fully demonstrate the capability of our method, we focus on time series data with non-stationarity. Unlike most classification tasks where the time series is windowed to represent samples of different classes, here we are interested in long time series where the behaviour of the signals might change over time. More information on the experiment datasets, cohort selection and processing, are provided in the Appendix \ref{app:dataset}.

%%%%%%%%%%%%%%%%%%%%%%%%%%%%%%%%%%%%%%%
%%%%%%%%%%% Downstream Task %%%%%%%%%%%
%%%%%%%%%%%%%%%%%%%%%%%%%%%%%%%%%%%%%%%

\subsection{Improving Downstream Prediction Tasks}

General representations should capture the important information in the data and can therefore be leveraged for downstream prediction tasks by training simple predictors on them. This approach is commonly used for evaluating the quality of representations \citep{oord2018representation, franceschi2019unsupervised, fortuin2020gp}. For the Physionet dataset, we consider the mortality prediction task. As a supervised baseline, we train an end-to-end model that directly uses the time series measurements to predict the risk of in-hospital mortality. For all other baselines, a simple Recurrent Neural Network (RNN) is trained to predict the risk using the representations over time. For the Air Quality dataset, the defined task is to estimate the average daily rain. Similarly, a simple RNN model is used for training this predictor using the representations.
For our approach and C-DSVAE where the global and local representations are encoded separately, the global representations are concatenated with the output of the RNN model to estimate the downstream task. Appendix \ref{app:exp} provides more details on the architecture of the models used in our experiments.

Table \ref{tb:downstream} shows the performance of all models on the prediction tasks. 
For better comparison, we have baselines with different representation dimensionality. The results show that our method outperforms others for the ICU mortality prediction task, with even fewer representation dimensions, and comes second to C-DSVAE for daily rain estimation. For Physionet, we even perform closely to a fully-supervised model. GP-VAE performs better than the regular VAE, as it properly models the correlation between representations of samples over time. As we increase the dimensionality of the representations, this model improves albeit with higher complexity and less interpretability in the encodings. By decoupling the representations, our method achieves superior performance with smaller dimensionality. Lastly, we demonstrate that the counterfactual regularization substantially improves the performance of our method (as shown by the performance results of our method with no regularization).

%%%%%%%%%%%%%%%%%%%%%%%%%%%%%%%%%%%%%%%
%%%%%%%%%%%%% Subgroups  %%%%%%%%%%%%%%
%%%%%%%%%%%%%%%%%%%%%%%%%%%%%%%%%%%%%%%

\subsection{Subgroup Identification}

\begin{figure}
    \centering
    \includegraphics[scale=0.5]{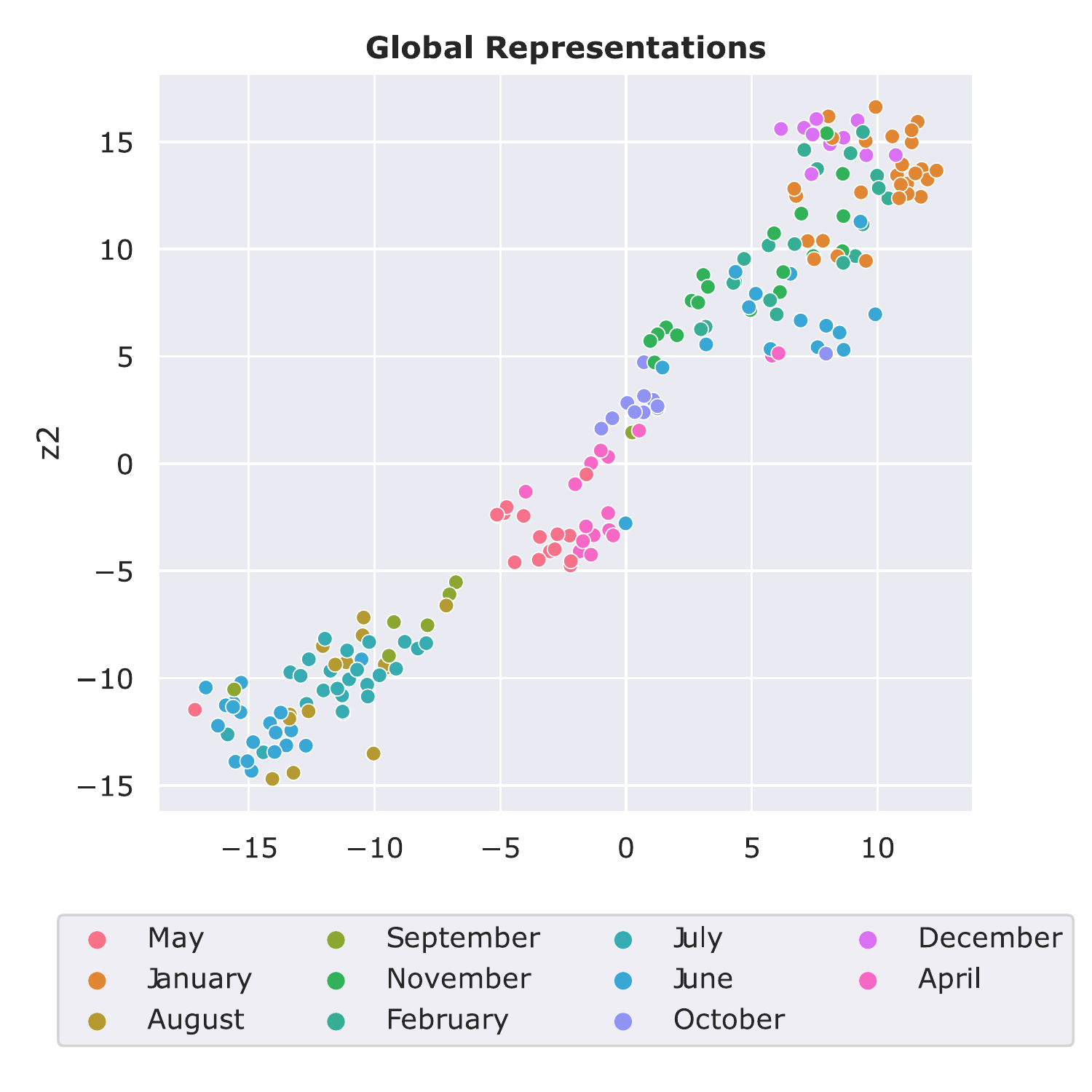}
    \caption{t-SNE visualization of the representations of the Air Quality dataset. Each data point is the global representation of a time series sample and the color indicates the month of year of that measurement.}
    \label{fig:global_reps}
\end{figure}

In many cases, we are interested in identifying or clustering samples with similar global properties invariant to other factors that can influence the time series trajectory. The global representations provide the information that allows us to identify such subgroups of the data. This is an important application, and as an example of the benefits, earlier work shows that in many applications, performing cluster-specific modelling or prediction can improve the overall performance of ML models \citep{giordano2019coherent, bouveyron2011model}. 
This experiment evaluates how well our global representations identify clusters of similar samples. For the Physionet dataset, we choose the ICU unit type as a proxy label for identifying sub groups 
% look into the unit type the patients were admitted to 
because patients that are admitted to these different units have some underlying similarities. For the Air Quality dataset, we are interested in identifying the month of the year that each recording belongs to.

\begin{table*}[t]
\begin{center}
\begin{tabular}{llcccc}
& & \multicolumn{2}{c}{Air Quality} & \multicolumn{2}{c}{Physionet}\\
\toprule
{\bf Model} & {\bf Dimensions}  & {\bf NLL} & {\bf MSE}& {\bf NLL} & {\bf MSE}
\\ \midrule
Our method & 8 $\times$ steps+8  & {\bf1.445} $\pm$ 0.052 & {\bf0.609} $\pm$ 0.026 & {\bf 2.609} $\pm$ 0.032 & {\bf 1.183} $\pm$ 0.016\\
Our method - no reg.  & 8 $\times$ steps+8 & 3.866 $\pm$ 0.344 & 0.906 $\pm$ 0.040 & 3.250 $\pm$ 0.145 & 1.323 $\pm$ 0.048\\
\midrule
GP-VAE & 8 $\times$ steps  & 1.908 $\pm$ 0.114 &  0.743 $\pm$ 0.048 &  2.626 $\pm$ 0.063 & 1.197 $\pm$ 0.031 \\
GP-VAE & 16 $\times$ steps  & 3.080 $\pm$ 0.281 &  0.941 $\pm$ 0.065 & 2.833 $\pm$ 0.069 & 1.295 $\pm$ 0.035\\
\bottomrule
\end{tabular}
\caption{Performance of different representation learning methods used for time series forecasting.}
\label{tb:forecasting}
\end{center}
\end{table*}

We use a simple MLP classifier to predict the subgroup of each sample using the learned representations. 
Our method and C-DSVAE define global representations separately, and that is what we use for the training. For the other baselines, we randomly select the representation of one window for this task as the representations encode both the local and global properties of the sample.
Table \ref{tab:global_classification} summarizes the classification performance of all baselines. Our results support the claim that the global representations capture the global characteristics of each sample. We can therefore identify samples with similar characteristics better than other baselines by only using these representations regardless of the changes over time. Fig. \ref{fig:global_reps} also shows the 2-dimensional projection of the global representations of the Air Quality dataset, learned using our method. We can observe the clear distinction between the global properties of measurements from the warmer months of the year and the ones around winter and fall. Even within these categories, the representations seem to cluster the months together.

%%%%%%%%%%%%%%%%%%%%%%%%%%%%%%%%%%%%%%%
%%%%%%%%%%%%% Forecasting %%%%%%%%%%%%%
%%%%%%%%%%%%%%%%%%%%%%%%%%%%%%%%%%%%%%%

\subsection{More Accurate Forecasting}

Forecasting high-dimensional time series is another important application in many domains, like retail and finance. Prior works observe that exploiting global patterns and coupling them with local calibration help prediction performance on many datasets \citep{sen2019think}. Our proposed method, designed in a similar vein for representation learning, can therefore be well-suited to improve forecasting performance. More importantly, since we model the local representations over time using a GP, the conditional distribution over the future local representations can be estimated by conditioning over the observed historical representations \citep{williams2006gaussian}; As shown in Eq. \ref{eq:forecast}. Here, $Z_l^*$ represent the local representations, and $t^*$ corresponds to the time steps in the future.

\begin{equation}
\small
\begin{aligned}
     Z_l^*\vline Z_l, t, t^* & \sim \mathcal{N}(\mu(Z_l^*), \Cov(Z_l^*))\\
    \Cov(Z_l^*) & = \mathbf{K}(Z_l^*, Z_l^*) - \mathbf{K}(Z_l^*, Z_l)\mathbf{K}(Z_l, Z_l)^{-1}\mathbf{K}(Z_l, Z_l^*)\\
     \mu(Z_l^*) & = \mathbf{K}(Z_l^*, Z_l)\mathbf{K}(Z_l, Z_l)^{-1}Z_l
\end{aligned}
\label{eq:forecast}
\end{equation}

We evaluate the performance of our method and the GP-VAE baselines for the time series forecasting, as shown by the results in Table \ref{tb:forecasting}. The objective is to predict two windows, equivalent to 28 observations or 2-days measurements for the Air Quality data and eight measurements in Physionet, using the full history of the signal. 
While deep learning models are prone to overfitting with increasing forecast horizon especially in non-stationary settings, our probabilistic method achieves good performance in such scenarios by estimating the expected trajectory of the signal using predictions for future local representations.
Additional plots showing the forecasting performance are provided in the Appendix \ref{app:extra_plot}.

%%%%%%%%%%%%%%%%%%%%%%%%%%%%%%%%%%%%%%%%
%%%%%%%%%%% Counterfactual %%%%%%%%%%%%%
%%%%%%%%%%%%%%%%%%%%%%%%%%%%%%%%%%%%%%%%

\subsection{Learning Disentangled Representations}

\begin{figure*}[!h]
    \centering
    \includegraphics[scale=0.7]{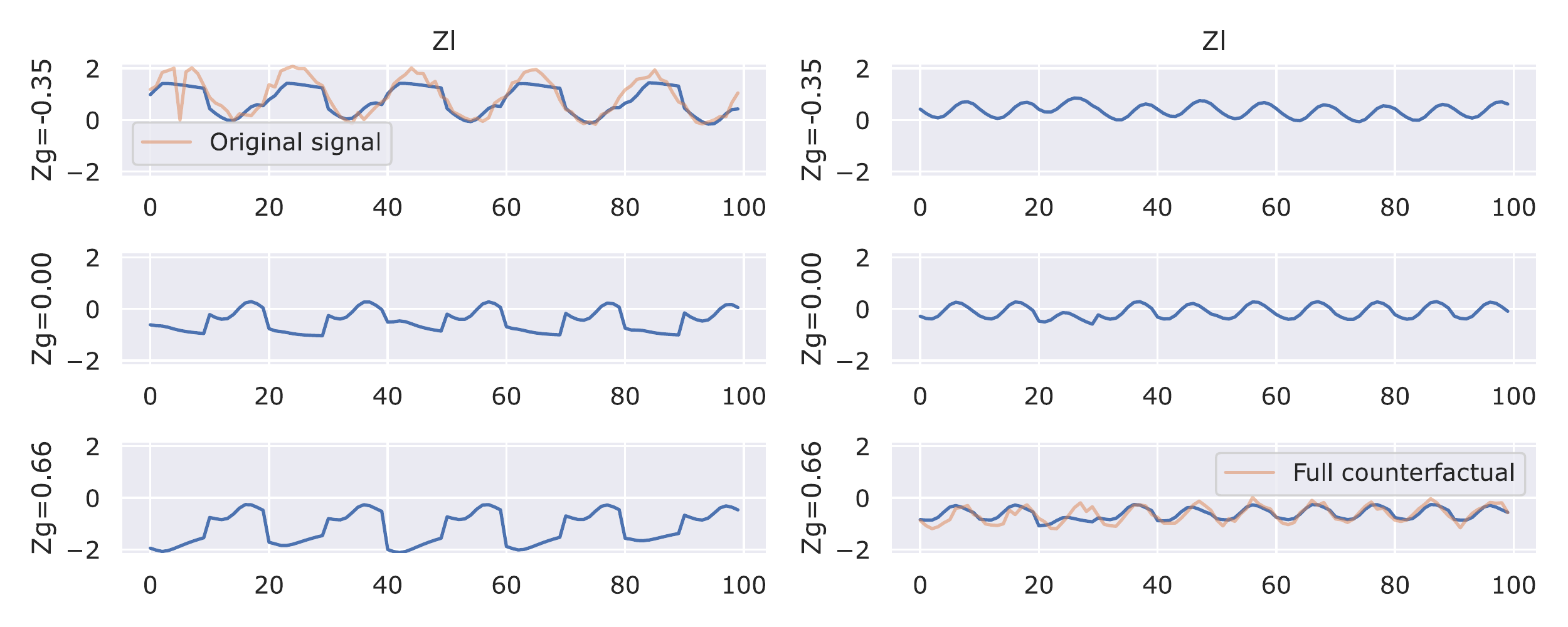}
    \caption{Examples of counterfactual signal generation in simulated dataset. The top left plot shows the original time-series in orange. Each row shows generated samples with the same underlying local representations as the original signal, with variable $\zg$. Each column represent different $\zl$ for the different $\zg$ values. }
    \label{fig:counterfactual_dist}
\end{figure*}

Many applications require generative models that are capable of synthesizing new instances where certain key factors of variation are held constant and other varied \citep{mathieu2016disentangling}. For example, in medical diagnosis, one may wish to model a treatment trajectory on different individuals. Decoupling the global representations from local variations enables such control over the generative process. We have curated a simulated dataset to test this functionality as the evaluation of generated counterfactual samples is difficult on real-world data without the knowledge of the ground truth. This dataset consists of time series samples composed of seasonality, trend, and noise. There are four types of signals in the dataset, with two global classes that determine the trend and intercept and two local classes that determine the seasonal component's frequency. More information on this dataset is provided in the Appendix \ref{app:dataset}.

Using our approach, we learn one-dimensional local and global representations for each sample.
We then generate new samples, controlling the local or global behaviours, and observe the effect of changes on the generated time-series. Fig. \ref{fig:counterfactual_dist} shows an example of this experiment. On the top left, we show a sample from the dataset. Over the multiple rows, we generate samples by gradually changing the global representation while keeping the local behaviour the same. We see that this changes the trend and intercept of the signal while maintaining the frequency. This is expected given our knowledge of the generative process of this data. Then, we keep $\zg$ constant for each sample and generate samples by changing the local representation in the second column. The change in local representations manifests as a change in the frequency, confirming our hypothesis. 
Note that some of the samples generated (Fig. \ref{fig:counterfactual_dist} middle row) are the ones never seen by the model, but still, their properties are correctly inferred. This experiment shows that we can learn the true underlying factors of data and shows that our generative approach can synthesize realistic samples by changing certain underlying factors of variation.

%% file: conclusion.tex
\section{Conclusion}

This paper introduces a generative approach for learning global and local representations for time series. We demonstrate the benefits of decoupling these representations for improving downstream performance, efficiency and a better understanding of the underlying generative factors of the data. 
Decoupling the underlying factors of variation in time series brings us one step closer to a better understanding of the generative process of complex temporal data. As a future direction, we would like to investigate how to associate the representation with concrete notions in the data space for improved interpretability and explore potential applications of counterfactual generation in real-world data.

\vspace{5mm}

%% file: appendix.tex
\section{SUPPLEMENTARY MATERIAL}
% \section{appendix}
\subsection{Datasets} \label{app:dataset}
This section provides additional details about each of the datasets used in our experiments.

\subsubsection{Physionet Dataset}
This dataset was made available as part of the PhysioNet Computing in Cardiology Challenge 2012 \footnote{\url{https://physionet.org/content/challenge-2012/1.0.0/}} with the objective of predicting mortality in ICU patients. The data consists of records from 12,000 ICU stays. All patients were adults who were admitted for a wide variety of reasons to cardiac, medical, surgical, and trauma ICUs. 
For cohort selection, out of the entire set of features that include multiple lab measurements over time and physiological signals collected at the bedside, we have selected the ones with less than 60 percent missing observations over the entire dataset. The length of all samples is restricted to be between 40 to 80 measurements. Some of the time series processing steps are borrowed from \citet{johnson2012patient} \footnote{\url{https://github.com/alistairewj/challenge2012}} (without any feature extraction for signals).

For all the experiments, we only use the time series measures for training the unsupervised representation learning frameworks, without the static information about each patient. We use the general descriptors as proxy labels for evaluating global representations throughout the experiments.

\subsubsection{Air Quality Dataset}
The Beijing multi-site Air Quality dataset \footnote{\url{https://archive.ics.uci.edu/ml/datasets/Beijing+Multi-Site+Air-Quality+Data}} includes hourly air pollutants data from 12 nationally-controlled air-quality monitoring sites, collected from the Beijing Municipal Environmental Monitoring Center. The meteorological data in each air-quality site are matched with the nearest weather station from the China Meteorological Administration. The time period is from March 1st, 2013, to February 28th, 2017. We create our dataset by dividing this time series into samples from different stations and of different months of the year. For our experiments, we use the pollutant measurements as our time series, the daily rain amount as a proxy for local behaviour and the station and month of each sample as the global characteristics. 

\subsubsection{Simulated Dataset}
We have created the simulated dataset to assess the quality of the counterfactual generated samples, as the ground-truth generative process of the data is known to us. The dataset consists of a total of 500 samples, with 100 measurements over time. Each sample is a one-dimensional time series and can be decomposed as follows: 

\begin{equation}
    \X_{(t)} = \alpha \times (\gamma t + a\sin(\frac{bt}{2\pi}) + c) + \epsilon
\end{equation}

Where $\epsilon \sim \mathcal{N}(0,0.1)$ is the Gaussian noise, and the rest of the parameters are determined by the global and local representation class of the sample. There are two possible classes defined for each of the local and global representations, resulting in overall four different types of time series samples in the cohort. 
Table \ref{tab:sim_data} describes the underlying generative process of each sample based on the underlying global and local variable. 

\begin{table}[]
    \centering
    \begin{tabular}{c|cc|cc}
         & \multicolumn{2}{c|}{$\zl$ = 1} & \multicolumn{2}{|c}{$\zl$ = 2}\\
        \hline
        $\zg$ = 1   & $\gamma=0.05$ & $a=1.8$   & $\gamma=0.05$ & $a=0.8$\\
                    & $c=-1.5$      &  $b=40$   & $c=-1.5$      & $b=20$\\
                    & \multicolumn{2}{c|}{$\alpha=0.5$} & \multicolumn{2}{c}{$\alpha=0.8$}\\
        \hline
        $\zg$ = 2   & $\gamma=-0.05$ & $a=1.8$  & $\gamma=-0.05$& $a=0.8$\\
                        & $c=+1.5$   &  $b=40$  & $c=+1.5$      & $b=20$\\
                    & \multicolumn{2}{c|}{$\alpha=0.5$} & \multicolumn{2}{c}{$\alpha=0.8$}\\
        % \hline
    \end{tabular}
    \caption{Parameters of the simulated dataset}
    \label{tab:sim_data}
\end{table}

\subsection{Experiment Setup} \label{app:exp}
For reproducibility purposes, the implementation code for all the experiments is made available. We have also provided additional information about each experiment in this section.

\subsubsection{Training our model}

For training our method on all the different datasets, we have chosen the list of hyper-parameters, reported in Table \ref{tab:hyper_param}. The window and the representation sizes are a design choice and depend on the properties of the data The rest of the parameters are selected based on the model performance on the validation set.

\begin{table*}[!h]
    \centering
    \begin{tabular}{cccc}
    Parameter & Physionet & Air Quality & Simulation\\
    \toprule
    $\lambda$   &  1.0 & 0.5 & 2.0\\
     $\beta$    & 0.1 & 0.1 & 0.1\\
     Window size ($\delta$)   & 4 & 24 & 10\\
     Global representation size ($d_g$)  & 8 & 8 & 1\\
     Local representation size ($d_l$)  & 8 & 8 & 1\\
     Optimizer  & Adam & Adam & Adam\\
     Learning rate  & 0.001 & 0.001 & 0.01\\
     Prior Kernels  & RBF, Cauchy & RBF, Cauchy & RBF\\
     Prior Kernel scales  & 2, 1, 0.5, 0.25 & 2, 1, 0.5, 0.25 & 1\\
     
     \bottomrule
    \end{tabular}
    \caption{List of the selected parameters for training our method on different dataset}
    \label{tab:hyper_param}
\end{table*}

\subsubsection{Downstream task}
The downstream task test evaluates the generalizability and usability of the learned representations. All baseline methods learn a representation for each window of time series. In addition to the representations over time, our method learns a single representation vector for the global variable. For the mortality prediction task, we use a single layer RNN, followed by a fully connected layer to estimate the risk of mortality. To integrate the global representation, our method concatenates the final hidden state of the RNN with the global representation vector. 
The designated task for the Air Quality dataset is the prediction of average daily rain. Each local representation encodes a window of 24 samples, equivalent to a day of measurements. A 2-layer MLP uses the local representations concatenated with the global representation to predict this value. A big challenge with the rain prediction is that it is a highly imbalanced regression problem, where on almost $90\%$ of the days, there is no rain, and for the rest, the amount is highly variable. To mitigate this, we use a weighted mean absolute error loss for training, and report the mean squared error as a performance metric to ensure consistency among results.

\subsubsection{Subgroup classification}
For this test, the goal is to identify similar subgroups using the global representations. For baselines where the two representations are not separated, we randomly select one of the local representations over time to use for the prediction. The global properties are expected to be encoded in the same representation vector in these baselines. Using a two layer MLP and the learned representations, we classify samples into the subgroups defined by out proxy labels (type of ICU for Physionet and month of year for the Air quality dataset).

\subsection{Supplementary Plots} \label{app:extra_plot}
% This section provides the supplementary figure of the paper.

\begin{figure}[!h]
    \centering
    \includegraphics[scale=0.35]{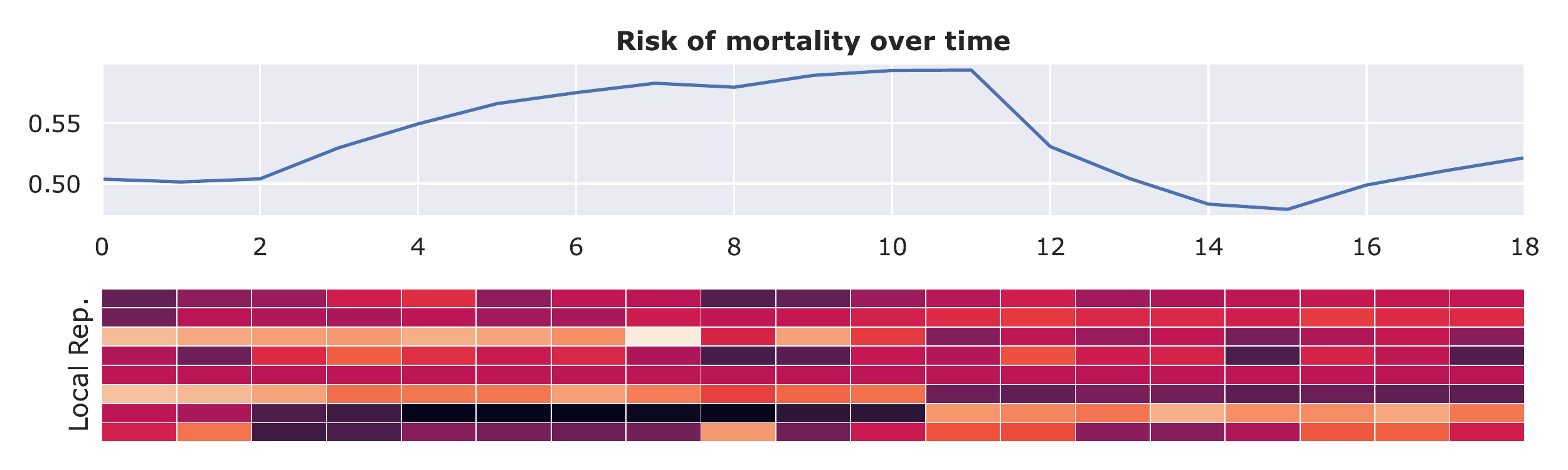}
    \caption{Risk of mortality estimation over time, based on the local and global representations (As part of the downstream task test for Physionet data). }
    \label{fig:mortality_risk}
\end{figure}

\begin{figure}
    \centering
    \includegraphics[scale=0.35]{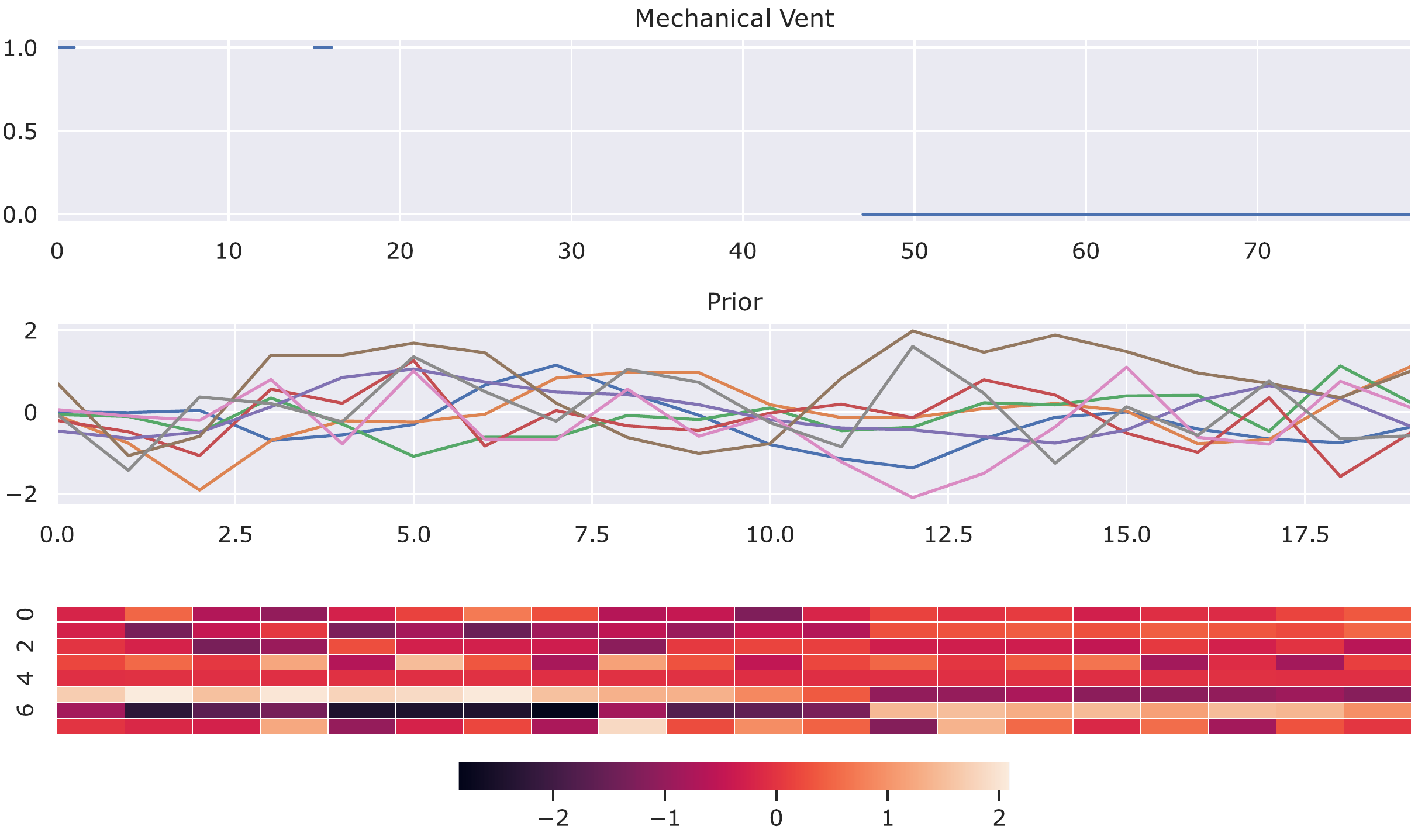}
    \caption{Exploratory analysis of local representations in Physionet data. The bottom heatmap demonstrates the 8 dimensional local representation of windows over time (x-axis). The middle bar shows the priors over all the different dimensions and finally the plot on the top shows the indicator of the mechanical ventillation. In the representation patterns, we can see signals indicating whether a patient is ventilated or not. Note that the ventilation information is never provided as an input to any of the encoders.}
    \label{fig:my_label}
\end{figure}

\begin{figure*}
    \centering
    \includegraphics[scale=0.5]{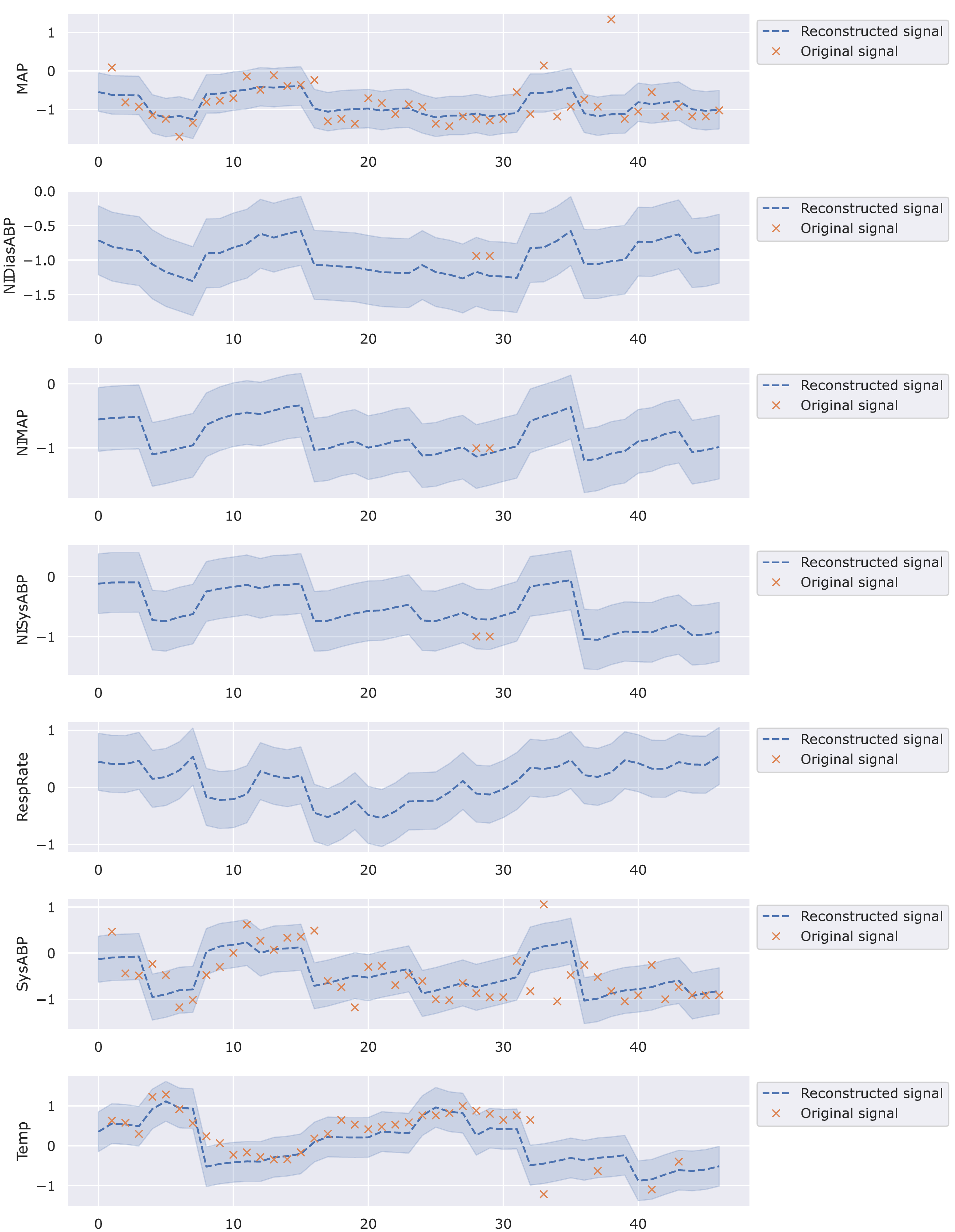}
    \caption{Reconstructed of a Physionet sample with high missing rate}
    \label{fig:physionet_reconstruction}
\end{figure*}

\begin{figure*}
    \centering
    \includegraphics[scale=0.4]{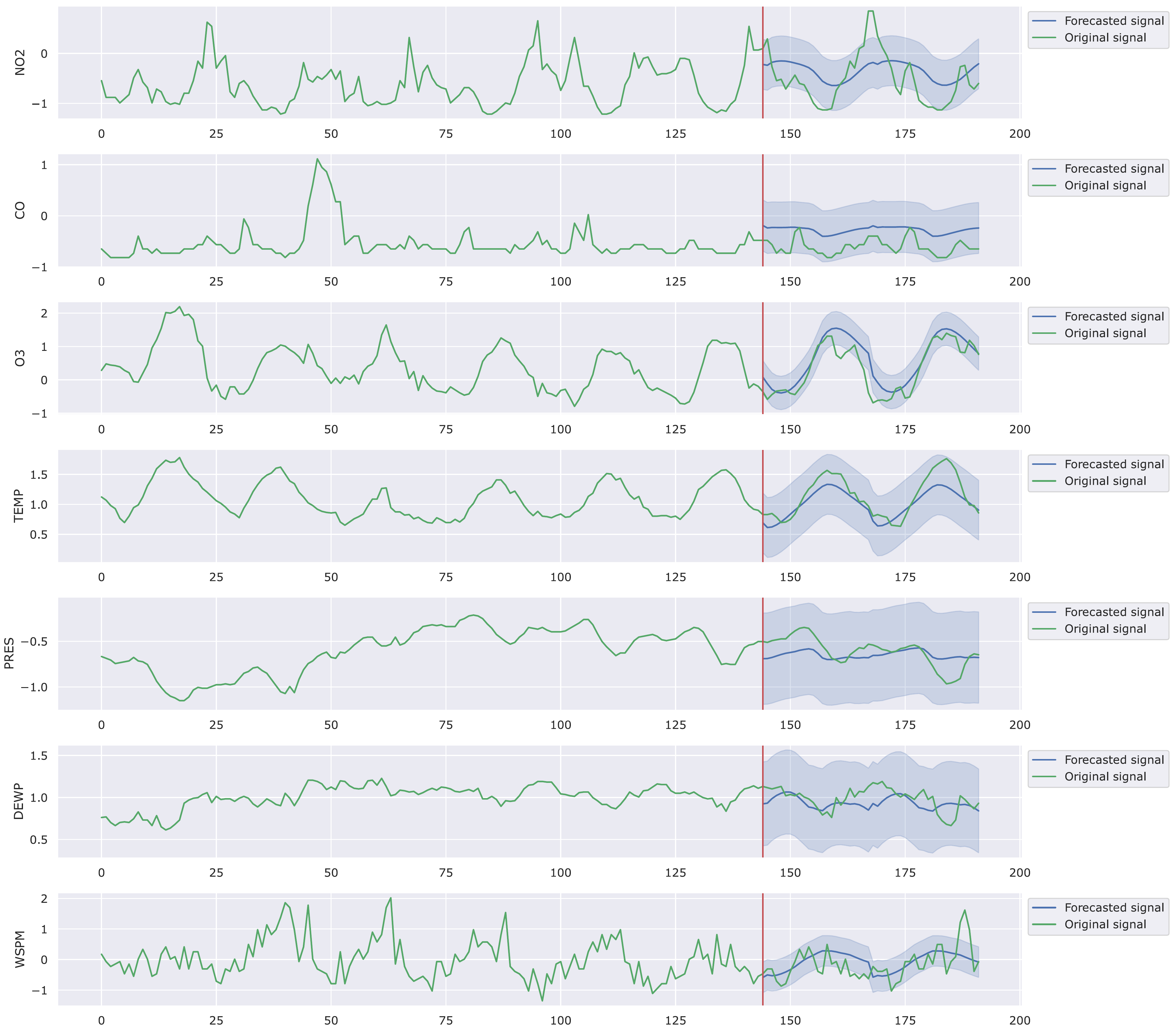}
    \caption{Forecasting pollutant measurements for the Air Quality dataset. The measurements before the vertical red line are observed (shown by the green line), and the forecasted measurements for 2 windows of time series are shown with the blue line, with the green plot demonstrating the expected prediction. The shaded regions indicate one standard deviation around the estimated distribution. }
    \label{fig:forecast_aq}
\end{figure*}

%% file: main.bbl
\begin{thebibliography}{}

\bibitem[Bai et~al., 2021]{bai2021contrastively}
Bai, J., Wang, W., and Gomes, C.~P. (2021).
\newblock Contrastively disentangled sequential variational autoencoder.
\newblock {\em NeurIPS}, 34.

\bibitem[Bai et~al., 2018]{bai2018interpretable}
Bai, T., Zhang, S., Egleston, B.~L., and Vucetic, S. (2018).
\newblock Interpretable representation learning for healthcare via capturing
  disease progression through time.
\newblock In {\em KDD}.

\bibitem[Bengio et~al., 2013]{bengio2013representation}
Bengio, Y., Courville, A., and Vincent, P. (2013).
\newblock Representation learning: A review and new perspectives.
\newblock {\em PAMI}, 35(8):1798--1828.

\bibitem[Bengio and LeCun, 2007]{Bengio+chapter2007}
Bengio, Y. and LeCun, Y. (2007).
\newblock Scaling learning algorithms towards {AI}.
\newblock In {\em Large Scale Kernel Machines}. MIT Press.

\bibitem[Bouveyron and Jacques, 2011]{bouveyron2011model}
Bouveyron, C. and Jacques, J. (2011).
\newblock Model-based clustering of time series in group-specific functional
  subspaces.
\newblock {\em Advances in Data Analysis and Classification}, 5(4):281--300.

\bibitem[Casale et~al., 2018]{casale2018gaussian}
Casale, F.~P., Dalca, A.~V., Saglietti, L., Listgarten, J., and Fusi, N.
  (2018).
\newblock Gaussian process prior variational autoencoders.
\newblock In {\em NeurIPS}.

\bibitem[Chen et~al., 2012]{chen2012marginalized}
Chen, M., Xu, Z., Weinberger, K.~Q., and Sha, F. (2012).
\newblock Marginalized denoising autoencoders for domain adaptation.
\newblock In {\em ICML}.

\bibitem[Chen et~al., 2016]{chen2016infogan}
Chen, X., Duan, Y., Houthooft, R., Schulman, J., Sutskever, I., and Abbeel, P.
  (2016).
\newblock Infogan: Interpretable representation learning by information
  maximizing generative adversarial nets.
\newblock In {\em NIPS}, pages 2180--2188.

\bibitem[Cheng et~al., 2020]{cheng2020club}
Cheng, P., Hao, W., Dai, S., Liu, J., Gan, Z., and Carin, L. (2020).
\newblock Club: A contrastive log-ratio upper bound of mutual information.
\newblock In {\em ICML}.

\bibitem[Chorowski et~al., 2019]{chorowski2019unsupervised}
Chorowski, J., Weiss, R.~J., Bengio, S., and van~den Oord, A. (2019).
\newblock Unsupervised speech representation learning using wavenet
  autoencoders.
\newblock {\em IEEE Trans audio, speech, and language processing},
  27(12):2041--2053.

\bibitem[Dezfouli et~al., 2019]{dezfouli2019disentangled}
Dezfouli, A., Ashtiani, H., Ghattas, O., Nock, R., Dayan, P., and Ong, C.~S.
  (2019).
\newblock Disentangled behavioral representations.
\newblock In {\em NeurIPS}.

\bibitem[Fortuin et~al., 2020]{fortuin2020gp}
Fortuin, V., Baranchuk, D., R{\"a}tsch, G., and Mandt, S. (2020).
\newblock Gp-vae: Deep probabilistic time series imputation.
\newblock In {\em AISTATS}.

\bibitem[Fortuin et~al., 2018]{fortuin2018som}
Fortuin, V., H{\"u}ser, M., Locatello, F., Strathmann, H., and R{\"a}tsch, G.
  (2018).
\newblock Som-vae: Interpretable discrete representation learning on time
  series.
\newblock In {\em ICLR}.

\bibitem[Franceschi et~al., 2019]{franceschi2019unsupervised}
Franceschi, J.-Y., Dieuleveut, A., and Jaggi, M. (2019).
\newblock Unsupervised scalable representation learning for multivariate time
  series.
\newblock {\em NeurIPS}.

\bibitem[Ghahramani, 2015]{ghahramani2015probabilistic}
Ghahramani, Z. (2015).
\newblock Probabilistic machine learning and artificial intelligence.
\newblock {\em Nature}, 521(7553):452--459.

\bibitem[Giordano et~al., 2019]{giordano2019coherent}
Giordano, G., Haberman, S., and Russolillo, M. (2019).
\newblock Coherent modeling of mortality patterns for age-specific subgroups.
\newblock {\em Decisions in Economics and Finance}, 42(1):189--204.

\bibitem[Goldberger et~al., 2000]{goldberger2000physiobank}
Goldberger, A.~L., Amaral, L.~A., Glass, L., Hausdorff, J.~M., Ivanov, P.~C.,
  Mark, R.~G., Mietus, J.~E., Moody, G.~B., Peng, C.-K., and Stanley, H.~E.
  (2000).
\newblock Physiobank, physiotoolkit, and physionet: components of a new
  research resource for complex physiologic signals.
\newblock {\em Circulation}, 101(23):e215--e220.

\bibitem[Hsu et~al., 2017]{hsu2017unsupervised}
Hsu, W.-N., Zhang, Y., and Glass, J. (2017).
\newblock Unsupervised learning of disentangled and interpretable
  representations from sequential data.
\newblock {\em arXiv preprint arXiv:1709.07902}.

\bibitem[Hyvarinen and Morioka, 2016]{hyvarinen2016unsupervised}
Hyvarinen, A. and Morioka, H. (2016).
\newblock Unsupervised feature extraction by time-contrastive learning and
  nonlinear ica.
\newblock {\em NIPS}.

\bibitem[Hyvarinen et~al., 2019]{hyvarinen2019nonlinear}
Hyvarinen, A., Sasaki, H., and Turner, R. (2019).
\newblock Nonlinear ica using auxiliary variables and generalized contrastive
  learning.
\newblock In {\em AISTATS}.

\bibitem[Johnson et~al., 2012]{johnson2012patient}
Johnson, A.~E., Dunkley, N., Mayaud, L., Tsanas, A., Kramer, A.~A., and
  Clifford, G.~D. (2012).
\newblock Patient specific predictions in the intensive care unit using a
  bayesian ensemble.
\newblock In {\em Computing in Cardiology (CinC), 2012}, pages 249--252. IEEE.

\bibitem[Kim and Mnih, 2018]{kim2018disentangling}
Kim, H. and Mnih, A. (2018).
\newblock Disentangling by factorising.
\newblock In {\em ICML}.

\bibitem[Klys et~al., 2018]{klys2018learning}
Klys, J., Snell, J., and Zemel, R. (2018).
\newblock Learning latent subspaces in variational autoencoders.
\newblock {\em arXiv preprint arXiv:1812.06190}.

\bibitem[Kumar et~al., 2018]{kumar2018variational}
Kumar, A., Sattigeri, P., and Balakrishnan, A. (2018).
\newblock Variational inference of disentangled latent concepts from unlabeled
  observations.
\newblock In {\em ICLR}.

\bibitem[L{\"a}ngkvist et~al., 2014]{langkvist2014review}
L{\"a}ngkvist, M., Karlsson, L., and Loutfi, A. (2014).
\newblock A review of unsupervised feature learning and deep learning for
  time-series modeling.
\newblock {\em Pattern Recognition Letters}, 42:11--24.

\bibitem[Lei et~al., 2019]{lei2019similarity}
Lei, Q., Yi, J., Vaculin, R., Wu, L., and Dhillon, I.~S. (2019).
\newblock Similarity preserving representation learning for time series
  clustering.
\newblock In {\em IJCAI}, volume~19, pages 2845--2851.

\bibitem[Ma et~al., 2019]{ma2019learning}
Ma, Q., Zheng, J., Li, S., and Cottrell, G.~W. (2019).
\newblock Learning representations for time series clustering.
\newblock {\em NeurIPS}.

\bibitem[Ma et~al., 2020]{ma2020decoupling}
Ma, X., Kong, X., Zhang, S., and Hovy, E.~H. (2020).
\newblock Decoupling global and local representations via invertible generative
  flows.
\newblock In {\em ICLR}.

\bibitem[Mallik, 2001]{mallik2001inverse}
Mallik, R.~K. (2001).
\newblock The inverse of a tridiagonal matrix.
\newblock {\em Linear Algebra and its Applications}, 325(1-3):109--139.

\bibitem[Mathieu et~al., 2016]{mathieu2016disentangling}
Mathieu, M.~F., Zhao, J.~J., Zhao, J., Ramesh, A., Sprechmann, P., and LeCun,
  Y. (2016).
\newblock Disentangling factors of variation in deep representation using
  adversarial training.
\newblock {\em NIPS}.

\bibitem[Moyer et~al., 2018]{moyer2018invariant}
Moyer, D., Gao, S., Brekelmans, R., Galstyan, A., and Ver~Steeg, G. (2018).
\newblock Invariant representations without adversarial training.
\newblock {\em NIPS}, 31:9084--9093.

\bibitem[Nguyen and Quanz, 2021]{nguyen2021temporal}
Nguyen, N. and Quanz, B. (2021).
\newblock Temporal latent auto-encoder: A method for probabilistic multivariate
  time series forecasting.
\newblock In {\em Proceedings of the AAAI Conference on Artificial
  Intelligence}, volume~35, pages 9117--9125.

\bibitem[Oord et~al., 2018]{oord2018representation}
Oord, A. v.~d., Li, Y., and Vinyals, O. (2018).
\newblock Representation learning with contrastive predictive coding.
\newblock {\em arXiv preprint arXiv:1807.03748}.

\bibitem[Reed et~al., 2015]{reed2015deep}
Reed, S.~E., Zhang, Y., Zhang, Y., and Lee, H. (2015).
\newblock Deep visual analogy-making.
\newblock {\em NIPS}.

\bibitem[Roberts et~al., 2013]{roberts2013gaussian}
Roberts, S., Osborne, M., Ebden, M., Reece, S., Gibson, N., and Aigrain, S.
  (2013).
\newblock Gaussian processes for time-series modelling.
\newblock {\em Philosophical Transactions of the Royal Society A: Mathematical,
  Physical and Engineering Sciences}, 371(1984):20110550.

\bibitem[Schulam and Saria, 2015]{schulam2015framework}
Schulam, P. and Saria, S. (2015).
\newblock A framework for individualizing predictions of disease trajectories
  by exploiting multi-resolution structure.
\newblock {\em NIPS}.

\bibitem[Sen et~al., 2019]{sen2019think}
Sen, R., Yu, H.-F., and Dhillon, I. (2019).
\newblock Think globally, act locally: A deep neural network approach to
  high-dimensional time series forecasting.
\newblock {\em arXiv preprint arXiv:1905.03806}.

\bibitem[Tenenbaum and Freeman, 2000]{tenenbaum2000separating}
Tenenbaum, J.~B. and Freeman, W.~T. (2000).
\newblock Separating style and content with bilinear models.
\newblock {\em Neural computation}, 12(6):1247--1283.

\bibitem[Tonekaboni et~al., 2020]{tonekaboni2020unsupervised}
Tonekaboni, S., Eytan, D., and Goldenberg, A. (2020).
\newblock Unsupervised representation learning for time series with temporal
  neighborhood coding.
\newblock In {\em ICLR}.

\bibitem[Wang et~al., 2019]{wang2019deep}
Wang, Y., Smola, A., Maddix, D., Gasthaus, J., Foster, D., and Januschowski, T.
  (2019).
\newblock Deep factors for forecasting.
\newblock In {\em International conference on machine learning}, pages
  6607--6617. PMLR.

\bibitem[Williams and Rasmussen, 2006]{williams2006gaussian}
Williams, C.~K. and Rasmussen, C.~E. (2006).
\newblock {\em Gaussian processes for machine learning}, volume~2.
\newblock MIT press Cambridge, MA.

\bibitem[Yang et~al., 2015]{yang2015weakly}
Yang, J., Reed, S.~E., Yang, M.-H., and Lee, H. (2015).
\newblock Weakly-supervised disentangling with recurrent transformations for 3d
  view synthesis.
\newblock In {\em NIPS}.

\bibitem[Yang and Wu, 2006]{yang200610}
Yang, Q. and Wu, X. (2006).
\newblock 10 challenging problems in data mining research.
\newblock {\em International Journal of Information Technology \& Decision
  Making}, 5(04):597--604.

\bibitem[Yingzhen and Mandt, 2018]{yingzhen2018disentangled}
Yingzhen, L. and Mandt, S. (2018).
\newblock Disentangled sequential autoencoder.
\newblock In {\em International Conference on Machine Learning}, pages
  5670--5679. PMLR.

\bibitem[Yuan et~al., 2019]{yuan2019wave2vec}
Yuan, Y., Xun, G., Suo, Q., Jia, K., and Zhang, A. (2019).
\newblock Wave2vec: Deep representation learning for clinical temporal data.
\newblock {\em Neurocomputing}, 324:31--42.

\bibitem[Zhang et~al., 2017]{zhang2017cautionary}
Zhang, S., Guo, B., Dong, A., He, J., Xu, Z., and Chen, S.~X. (2017).
\newblock Cautionary tales on air-quality improvement in beijing.
\newblock {\em Proceedings of the Royal Society A: Mathematical, Physical and
  Engineering Sciences}, 473(2205):20170457.

\bibitem[Zhu et~al., 2020]{zhu2020s3vae}
Zhu, Y., Min, M.~R., Kadav, A., and Graf, H.~P. (2020).
\newblock S3vae: Self-supervised sequential vae for representation
  disentanglement and data generation.
\newblock In {\em Proceedings of the IEEE/CVF Conference on Computer Vision and
  Pattern Recognition}, pages 6538--6547.

\end{thebibliography}
